%% file: main.tex
\documentclass[10pt,twocolumn,letterpaper]{article}
\newif\ifcomments
\commentsfalse
\commentstrue

\usepackage{cvpr}
\usepackage{times}
\usepackage{epsfig}
\usepackage{graphicx}
\usepackage{amsmath}
\usepackage{amssymb}
\usepackage{wrapfig}
\usepackage{multirow}
\usepackage[T1]{fontenc}
\usepackage[font=small,labelfont=bf,tableposition=top]{caption}

\DeclareCaptionLabelFormat{andtable}{#1~#2  \&  \tablename~\thetable}

\usepackage[pagebackref=true,breaklinks=true,letterpaper=true,colorlinks,bookmarks=false]{hyperref}

\cvprfinalcopy 

\def\cvprPaperID{1954} 

\ifcvprfinal\fi
\begin{document}

\ifcomments
  \newcommand{\comments}[1]{#1}
\else
  \newcommand{\comments}[1]{}
\fi

\newcommand{\name}{Deep Feature Interpolation}
\newcommand{\nameshort}{DFI}

\newcommand{\kqw}[1]{\comments{\textcolor{blue}{[Kilian: #1]}}}

\title{Densely Connected Convolutional Networks}

%

\author{Gao Huang\thanks{Authors contributed equally}\\
Cornell University\\
{\tt\small gh349@cornell.edu}
\and
Zhuang Liu$^*$\\
Tsinghua University\\
{\tt\small liuzhuang13@mails.tsinghua.edu.cn}
\and
Laurens van der Maaten\\
Facebook AI Research\\
{\tt\small lvdmaaten@fb.com}
\and
Kilian Q. Weinberger\\
Cornell University\\
{\tt\small kqw4@cornell.edu}
}

\input{macros}

\maketitle

\begin{abstract}
\input{abstract.tex}

\end{abstract}


\input{intro}

\input{related_work}

\input{method}
\input{experiments}

\input{analytic}

\input{conclusion}

\vspace{-3pt}
\paragraph{Acknowledgements.} The authors are supported in part by the NSF III-1618134, III-1526012, IIS-1149882, the Office of Naval Research Grant N00014-17-1-2175 and the Bill and Melinda Gates foundation. GH is supported by the International Postdoctoral Exchange Fellowship Program of China Postdoctoral Council (No.20150015). ZL is supported by the National Basic Research Program of China Grants 2011CBA00300, 2011CBA00301, the NSFC 61361136003. We also thank Daniel Sedra, Geoff Pleiss and Yu Sun
for many insightful discussions.

{\small
\bibliographystyle{ieee}
\bibliography{citations}}

%

\maketitle

\end{document}

%% file: macros.tex
\newcommand{\methodname}{dense convolutional  network}
\newcommand{\methodnamecap}{Dense Convolutional Network}
\newcommand{\methodnameshort}{DenseNet}
\newcommand{\methodnameshorts}{DenseNets}
\newcommand{\methodblock}{dense block}
\newcommand{\methodblockcap}{Dense Block}

\newcommand{\regmethodname}{feature drop}
\newcommand{\regmethodnamecap}{Feature Drop}

\newcommand{\stepsizename}{growth rate}

\newcommand{\conv}[1]{$\left[\begin{array}{ll} \text{1}\times \text{1} \text{ conv}\\ \text{3}\times \text{3} \text{ conv} \end{array}\right] \times \text{#1}$}

\newcommand{\cross}[1]{#1 $\times$ #1}

\newcommand{\feati}{x_i}
\newcommand{\clsfeati}{y_i}
\newcommand{\featk}{x_k}
\newcommand{\clsfeatk}{y_k}
\newcommand{\loss}{L}
\newcommand{\featL}{x_L}
\newcommand{\clsfeat}{y}
\newcommand{\anyxs}{\ensuremath{\mathbf{x}}}
\newcommand{\anyys}{\ensuremath{\mathbf{y}}}

\newcommand{\bx}{\ensuremath{\mathbf{x}}}

\newcommand{\sourcexs}{\ensuremath{\mathbf{x^\mathcal{S}}}}
\newcommand{\sourceys}{\ensuremath{\mathbf{y^\mathcal{S}}}}

\newcommand{\targetxs}{\ensuremath{\mathbf{x^\mathcal{T}}}}
\newcommand{\targetys}{\ensuremath{\mathbf{y^\mathcal{T}}}}
\newcommand{\pseudotargetys}{\ensuremath{\mathbf{\hat{y}^\mathcal{T}}}}

%% file: abstract.tex
Recent work has shown that convolutional networks can be substantially deeper, more accurate, and efficient to train if they contain shorter connections between layers close to the input and those close to the output.
In this paper, we embrace this observation and introduce the \methodnamecap{} (\methodnameshort{}), which connects each layer to every other layer in a feed-forward fashion. Whereas traditional convolutional networks with $L$ layers have $L$ connections---one between each layer and its subsequent layer---our network has $\frac{L(L+1)}{2}$ direct connections.
For each layer, the feature-maps of all preceding layers are used as inputs, and its own feature-maps are used as inputs into all subsequent layers.
\methodnameshorts{} have several compelling advantages:
they alleviate the vanishing-gradient problem, strengthen feature propagation, encourage feature reuse, and substantially reduce the number of parameters.
We evaluate our proposed architecture on four highly competitive object recognition benchmark tasks (CIFAR-10, CIFAR-100, SVHN, and ImageNet). \methodnameshorts{} obtain significant improvements over the state-of-the-art on most of them, whilst requiring less computation to achieve high performance. Code and pre-trained models are available at \url{https://github.com/liuzhuang13/DenseNet}.

%% file: intro.tex
\section{Introduction}

Convolutional neural networks (CNNs) have become the dominant machine learning approach for visual object recognition. Although they were originally introduced over 20 years ago~\cite{lecun}, improvements in computer hardware and network structure have enabled the training of  truly deep CNNs only recently. The original LeNet5~\cite{lenet5} consisted of 5 layers, VGG featured 19~\cite{vgg}, and only last year Highway Networks~\cite{highway} and Residual Networks (ResNets)~\cite{resnet} have surpassed the 100-layer barrier.

\begin{figure}[t]
      \centering
      \includegraphics[width=0.5\textwidth]{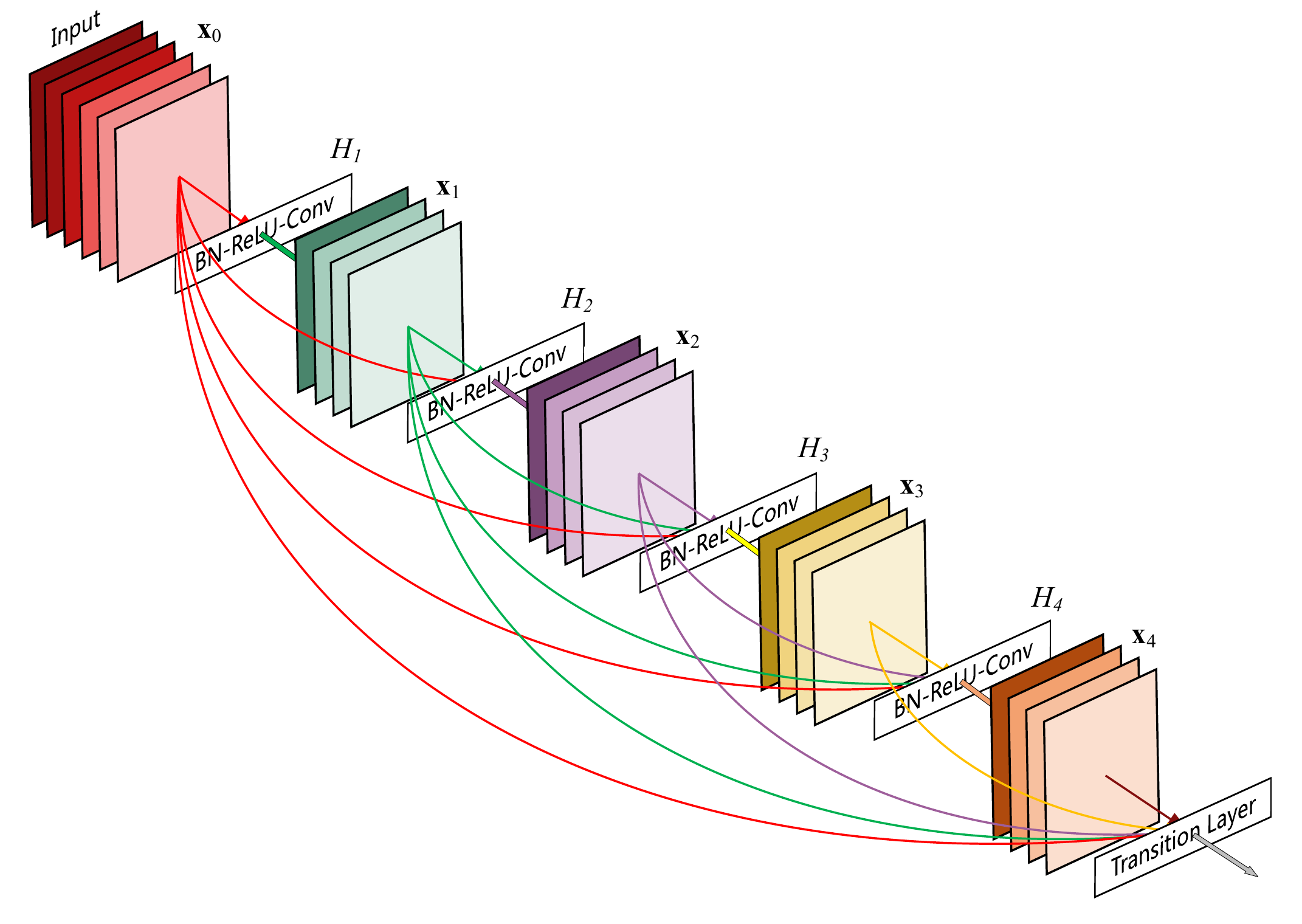}
      \caption{A 5-layer \methodblock{} with a growth rate of $k=4$. Each layer takes all preceding feature-maps as input. }
      \label{fig:ccnn}
\end{figure}

As CNNs become increasingly deep, a new research problem emerges: as information about the input or gradient passes through many layers, it can vanish and ``wash out'' by the time it reaches the end (or beginning) of the network.
Many recent publications address this or related problems. 
ResNets~\cite{resnet} and Highway Networks~\cite{highway} bypass signal from one layer to the next via identity connections. Stochastic depth~\cite{stochastic} shortens ResNets by randomly dropping layers during training to allow better information and gradient flow.
FractalNets \cite{fractalnet} repeatedly combine several parallel layer sequences with different number of convolutional blocks to obtain a large nominal depth, while maintaining many short paths in the network.
Although these different approaches vary in network topology and training procedure, they all share a key characteristic: they create short paths from early layers to later layers.

In this paper, we propose an architecture that distills this insight into a simple connectivity pattern: to ensure maximum information flow between layers in the network, we connect \emph{all layers} (with matching feature-map sizes) directly with each other. To preserve the feed-forward nature, each layer obtains additional inputs from all preceding layers and passes on its own feature-maps to all subsequent layers.
Figure~\ref{fig:ccnn} illustrates this layout schematically.
Crucially, in contrast to ResNets, we never combine features through summation before they are passed into a layer; instead, we combine features by concatenating them.
Hence, the $\ell^{th}$ layer has $\ell$ inputs, consisting of the feature-maps of all preceding convolutional blocks. Its own feature-maps are passed on to all $L-\ell$ subsequent layers. This introduces $\frac{L(L+1)}{2}$  connections in an $L$-layer network, instead of just $L$, as in traditional architectures.
Because of its dense connectivity pattern, we refer to our approach as {\emph{\methodnamecap{} (\methodnameshort{})}.

A possibly counter-intuitive effect of this dense connectivity pattern is that it requires \emph{fewer} parameters than traditional convolutional networks, as there is no need to re-learn redundant feature-maps.
Traditional feed-forward architectures can be viewed as algorithms with a state, which is passed on from layer to layer. Each layer reads the state from its preceding layer and writes to the subsequent layer. It changes the state but also passes on information that needs to be preserved. ResNets~\cite{resnet} make this information preservation explicit through additive identity transformations.
Recent variations of ResNets~\cite{stochastic} show that many layers contribute very little and can in fact be randomly dropped during training. This makes the state of ResNets similar to (unrolled) recurrent neural networks~\cite{liao2016bridging}, but the number of parameters of ResNets is substantially larger because each layer has its own weights.
Our proposed \methodnameshort{} architecture explicitly differentiates between information that is added to the network and information that is preserved.
\methodnameshort{} layers are very narrow (\emph{e.g.}, 12 filters per layer), adding only a small set of feature-maps to the ``collective knowledge'' of the network and keep the remaining feature-maps unchanged---and the final classifier makes a decision based on all feature-maps in the network.



Besides better parameter efficiency, one big advantage of \methodnameshort{}s is their improved flow of information and gradients throughout the network, which makes them easy to train. Each layer has direct access to the gradients from the loss function and the original input signal, leading to an implicit deep supervision \cite{dsn}. This helps training of deeper network architectures. Further, we also observe that dense connections have a regularizing effect, which reduces overfitting on tasks with smaller training set sizes.

We evaluate \methodnameshort{}s on four highly competitive benchmark datasets (CIFAR-10, CIFAR-100, SVHN, and ImageNet). Our models tend to require much fewer parameters than existing algorithms with comparable accuracy. Further, we significantly outperform the current state-of-the-art results on most of the benchmark tasks.

%% file: related_work.tex

\begin{figure*}[t]
\vspace{-2 ex}
      \centering
      \includegraphics[width=\textwidth]{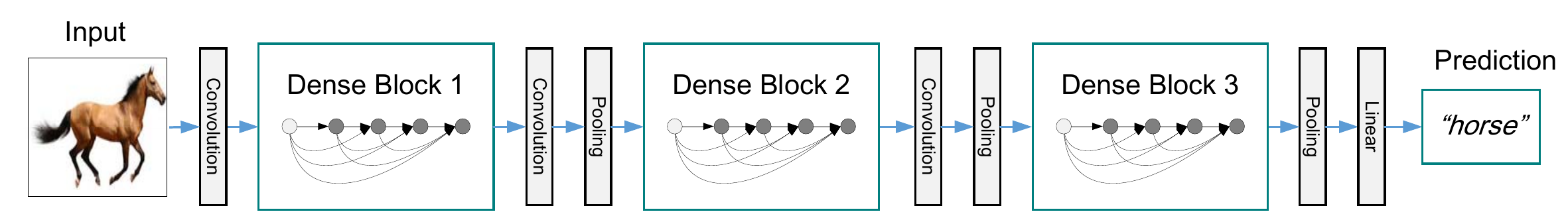}
      \caption{A deep \methodnameshort{} with three dense blocks. The layers between two adjacent blocks are referred to as transition layers and change feature-map sizes via convolution and pooling. }
      \label{fig:ccnn_all}
      \vspace{-2 ex}
\end{figure*}

\section{Related Work}
The exploration of network architectures has been a part of neural network research since their initial discovery.
The recent resurgence in popularity of neural networks has also revived this research domain.
The increasing number of layers in modern networks amplifies the differences between architectures and motivates the exploration of different connectivity patterns and the revisiting of old research ideas.

A {cascade structure} similar to our proposed dense network layout has already been studied in the neural networks literature in the 1980s \cite{fahlman1989cascade}. Their pioneering work focuses on fully connected multi-layer perceptrons trained in a layer-by-layer fashion. More recently, fully connected cascade networks to be trained with batch gradient descent were proposed \cite{wilamowski2010neural}. Although effective on small datasets, this approach only scales to networks with a few hundred parameters. 
In \cite{hypercolumn, fcn, pedestrian, yang2015multi}, utilizing multi-level features in CNNs through skip-connnections has been found to be effective for various vision tasks.
Parallel to our work, \cite{cortes2016adanet} derived a purely theoretical framework for networks with cross-layer connections similar to ours.

Highway Networks \cite{highway} were amongst the first architectures that provided a means to effectively train end-to-end networks with more than 100 layers. Using bypassing paths along with gating units, Highway Networks with hundreds of layers can be optimized without difficulty. The bypassing paths are presumed to be the key factor that eases the training of these very deep networks. This point is further supported by ResNets \cite{resnet}, in which pure identity mappings are used as bypassing paths. ResNets have achieved impressive, record-breaking performance on many challenging image recognition, localization, and detection tasks, such as ImageNet and COCO object detection~\cite{resnet}. Recently, \emph{stochastic depth} was proposed as a way to successfully train a 1202-layer ResNet \cite{stochastic}. Stochastic depth improves the training of deep residual networks by dropping layers randomly during training. This shows that not all layers may be needed and highlights that there is a great amount of redundancy in deep (residual) networks. Our paper was partly inspired by that observation. ResNets with \emph{pre-activation} also facilitate the training of state-of-the-art networks with $>$ 1000 layers \cite{identity-mappings}.

An orthogonal approach to making networks deeper (\emph{e.g.}, with the help of skip connections) is to increase the network \emph{width}. The GoogLeNet \cite{szegedy2015going,szegedy2015rethinking} uses an ``Inception module'' which concatenates feature-maps produced by filters of different sizes. In \cite{rir}, a variant of ResNets with wide generalized residual blocks was proposed. In fact, simply increasing the number of filters in each layer of ResNets can improve its performance provided the depth is sufficient \cite{wide}. FractalNets also achieve competitive results on several datasets using a wide network structure \cite{fractalnet}.

Instead of drawing representational power from extremely deep or wide architectures, \methodnameshorts{} exploit the potential of the network through \emph{feature reuse}, yielding condensed models that are easy to train and highly parameter-efficient. Concatenating feature-maps learned by \emph{different layers} increases variation in the input of subsequent layers and improves efficiency. This constitutes a major difference between \methodnameshorts{} and ResNets. Compared to Inception networks \cite{szegedy2015going,szegedy2015rethinking}, which also concatenate features from different layers, \methodnameshort{}s are simpler and more efficient.

There are other notable network architecture innovations which have yielded competitive results. The Network in Network (NIN) \cite{netinnet} structure includes micro multi-layer perceptrons into the filters of convolutional layers to extract more complicated features. In Deeply Supervised Network (DSN) \cite{dsn}, internal layers are directly supervised by auxiliary classifiers, which can strengthen the gradients received by earlier layers. Ladder Networks \cite{rasmus2015semi,pezeshki2015deconstructing} introduce lateral connections into autoencoders, producing impressive accuracies on semi-supervised learning tasks. In \cite{wang2016deeply}, Deeply-Fused Nets (DFNs) were proposed to improve information flow by combining intermediate layers of different base networks.
The augmentation of networks with pathways that minimize reconstruction losses was also shown to improve image classification models \cite{zhang2016augmenting}.

%
%
%
%
%
%

%% file: method.tex


\renewcommand{\arraystretch}{1.1}
\begin{table*}[!t]
\centering
\resizebox{0.85\textwidth}{!}{%
\begin{tabular}{c|c|c|l|l|l}
\hline
Layers                                                                          & Output Size     & DenseNet-121                                                                                                               & \multicolumn{1}{c|}{DenseNet-169}                                                                     & \multicolumn{1}{c|}{DenseNet-201}                                                                     & \multicolumn{1}{c}{DenseNet-264}                                                                       \\ \hline
Convolution                                                                     & \cross{112} & \multicolumn{4}{c}{\cross{7} conv, stride 2}                                                                                                                                                                                                                                                                                                                                                                                                    \\ \hline
Pooling                                                                         & \cross{56}   & \multicolumn{4}{c}{\cross{3} max pool, stride 2}                                                                                                                                                                                                                                                                                                                                                                                                \\ \hline
\begin{tabular}[c]{@{}c@{}}Dense Block\\ (1)\end{tabular}                       & \cross{56}   & \multicolumn{1}{l|}{\conv{6}}  & \conv{6}  & \conv{6}  & \conv{6} \\ \hline
\multirow{2}{*}{\begin{tabular}[c]{@{}c@{}}Transition Layer\\ (1)\end{tabular}} & \cross{56}  & \multicolumn{4}{c}{\cross{1} conv}                                                                                                                                                                                                                                                                                                                                                                                                              \\ \cline{2-6}
                                                                                & \cross{28}   & \multicolumn{4}{c}{\cross{2} average pool, stride 2}                                                                                                                                                                                                                                                                                                                                                                                             \\ \hline
\begin{tabular}[c]{@{}c@{}}Dense Block\\ (2)\end{tabular}                       & \cross{28}   & \multicolumn{1}{l|}{\conv{12}} & \conv{12}& \conv{12} & \conv{12} \\ \hline
\multirow{2}{*}{\begin{tabular}[c]{@{}c@{}}Transition Layer\\ (2)\end{tabular}} & \cross{28}  & \multicolumn{4}{c}{\cross{1} conv}                                                                                                                                                                                                                                                                                                                                                                                                              \\ \cline{2-6}
                                                                                & \cross{14}   & \multicolumn{4}{c}{\cross{2} average pool, stride 2}                                                                                                                                                                                                                                                                                                                                                                                             \\ \hline
\begin{tabular}[c]{@{}c@{}}Dense Block\\ (3)\end{tabular}                       & \cross{14}   & \multicolumn{1}{l|}{\conv{24}} & \conv{32} & \conv{48} & \conv{64} \\ \hline
\multirow{2}{*}{\begin{tabular}[c]{@{}c@{}}Transition Layer\\ (3)\end{tabular}} & \cross{14}   & \multicolumn{4}{c}{\cross{1} conv}                                                                                                                                                                                                                                                                                                                                                                                                              \\ \cline{2-6}
                                                                                & \cross{7}     & \multicolumn{4}{c}{\cross{2} average pool, stride 2}                                                                                                                                                                                                                                                                                                                                                                                             \\ \hline
\begin{tabular}[c]{@{}c@{}}Dense Block\\ (4)\end{tabular}                       & \cross{7}   & \multicolumn{1}{l|}{\conv{16}} & \conv{32} & \conv{32} & \conv{48}  \\ \hline
\multirow{2}{*}{\begin{tabular}[c]{@{}c@{}}Classification\\ Layer\end{tabular}} & \cross{1} & \multicolumn{4}{c}{\cross{7} global average pool}                                                                                                                                                                                                                                                                                                                                                                                            \\ \cline{2-6}
                                                                                &                 & \multicolumn{4}{c}{1000D fully-connected, softmax}                                                                                                                                                                                                                                                                                                                                                                                               \\ \hline
\end{tabular}
}
\vspace{1 ex}
\caption{DenseNet architectures for ImageNet. The growth rate for all the networks is $k=32$. Note that each ``conv'' layer shown in the table corresponds the sequence BN-ReLU-Conv.   }
\label{densenet-imagenet}
\vspace{-3 ex}
\end{table*}

\section{DenseNets}

Consider a single image $\bx_0$ that is passed through a convolutional network. The network comprises $L$ layers, each of which implements a non-linear transformation $H_\ell(\cdot)$, where $\ell$ indexes the layer. $H_\ell(\cdot)$ can be a composite function of operations such as Batch Normalization (BN) \cite{batch-norm}, rectified linear units (ReLU)~\cite{relu}, Pooling~\cite{lenet5}, or  Convolution (Conv).
We denote the output of the $\ell^{th}$ layer as $\bx_{\ell}$.

\vspace{-2 ex}
\paragraph{ResNets.}
Traditional convolutional feed-forward networks connect the output of the $\ell^{th}$ layer as input to the $(\ell+1)^{th}$ layer~\cite{alexnet}, which gives rise to the following layer transition: $\bx_\ell=H_\ell(\bx_{\ell-1})$. ResNets~\cite{resnet} add a skip-connection that bypasses the non-linear transformations with an identity function:
\vspace{-1ex}
\begin{equation}
\vspace{-1ex}
\bx_\ell=H_\ell(\bx_{\ell-1})+\bx_{\ell-1}\label{eq:resnet}.
\end{equation}
An advantage of ResNets is that the gradient can flow directly through the identity function from later layers to the earlier layers. However, the identity function and the output of $H_\ell$ are combined by summation, which may impede the information flow in the network.

\vspace{-2 ex}
\paragraph{Dense connectivity.}
To further improve the information flow between layers we propose a different connectivity pattern: we introduce direct connections from any layer to all subsequent layers. Figure~\ref{fig:ccnn} illustrates the layout of the resulting \methodnameshort{} schematically.
Consequently, the $\ell^{th}$ layer receives the feature-maps of all preceding layers, $\bx_0,\dots,\bx_{\ell-1}$, as input:
\vspace{-1ex}
\begin{equation}
\vspace{-1ex}
\bx_{\ell} = \ H_\ell([\bx_0, \bx_1,\ldots, \bx_{\ell-1}]),
\label{eqn:densenet}
\end{equation}
where $[\bx_0, \bx_1,\ldots, \bx_{\ell-1}]$ refers to the concatenation of the feature-maps produced in layers $0,\dots,\ell-1$. Because of its dense connectivity we refer to this network architecture as \emph{\methodnamecap{} (\methodnameshort{})}.
For ease of implementation, we concatenate the multiple inputs of $H_\ell(\cdot)$ in eq.~(\ref{eqn:densenet}) into a single tensor.

\vspace{-2 ex}
\paragraph{Composite function.}
Motivated by \cite{identity-mappings}, we define $H_\ell(\cdot)$ as a composite function of three consecutive operations: batch normalization (BN) \cite{batch-norm}, followed by a rectified linear unit (ReLU)~\cite{relu} and a $3\times 3$ convolution (Conv).


\vspace{-2 ex}
\paragraph{Pooling layers.}
The concatenation operation used in Eq.~(\ref{eqn:densenet}) is not viable when the size of feature-maps changes. However, an essential part of convolutional networks is down-sampling layers that change the size of feature-maps.  To facilitate down-sampling in our architecture we divide the network into multiple densely connected \emph{dense blocks}; see Figure~\ref{fig:ccnn_all}. We refer to layers between blocks as \emph{transition layers}, which do convolution and pooling. The transition layers used in our experiments consist of a batch normalization layer and an 1$\times$1 convolutional layer followed by a 2$\times$2 average pooling layer.


\vspace{-2 ex}
\paragraph{Growth rate.}
If each function $H_\ell$ produces $k$ feature-maps, it follows that the $\ell^{th}$ layer has $k_0+k\times (\ell-1)$ input feature-maps, where $k_0$ is the number of channels in the input layer. An important difference between \methodnameshort{} and existing network architectures is that \methodnameshort{} can have very narrow layers, \emph{e.g.}, $k=12$. We refer to the hyper-parameter $k$ as the \emph{\stepsizename{}} of the network. We show in Section~\ref{sec:results} that a relatively small \stepsizename{} is sufficient to obtain state-of-the-art results on the datasets that we tested on. One explanation for this is that each layer has access to all the preceding feature-maps in its block and, therefore, to the network's ``collective knowledge''. One can view the feature-maps as the global state of the network. Each layer adds $k$ feature-maps of its own to this state. The  \stepsizename{} regulates how much new information each layer contributes to the global state.
The global state, once written, can be accessed from everywhere within the network and, unlike in traditional network architectures, there is no need to replicate it from layer to layer.

\vspace{-2 ex}
\paragraph{Bottleneck layers.}
Although each layer only produces $k$ output feature-maps, it typically has many more inputs. It has been noted in \cite{szegedy2015rethinking,resnet} that a 1$\times$1 convolution can be introduced as \emph{bottleneck} layer before each 3$\times$3 convolution to reduce the number of input feature-maps, and thus to improve computational efficiency. We find this design especially effective for \methodnameshort{} and we refer to our network with such a bottleneck layer, \emph{i.e.}, to the BN-ReLU-Conv(1$\times$ 1)-BN-ReLU-Conv(3$\times$3) version of $H_\ell$, as \methodnameshort{}-B. In our experiments, we let each 1$\times$1 convolution produce $4k$ feature-maps.

\vspace{-2 ex}
\paragraph{Compression.}
To further improve model compactness, we can reduce the number of feature-maps at transition layers. If a dense block contains $m$ feature-maps, we let the following transition layer generate $\lfloor \theta m\rfloor$ output feature-maps, where $0<\!\theta\le \!1$ is referred to as the compression factor. When $\theta\!=\!1$, the number of feature-maps across transition layers remains unchanged. We refer the \methodnameshort{} with $\theta\!<\!1$ as DenseNet-C, and we set $\theta=0.5$ in our experiment. When both the bottleneck and transition layers with $\theta\!<1$ are used, we refer to our model as \methodnameshort{}-BC.

\vspace{-2 ex}
\paragraph{Implementation Details.}
On all datasets except ImageNet, the \methodnameshort{} used in our experiments has three dense blocks that each has an equal number of layers. Before entering the first dense block, a convolution with 16 (or twice the \stepsizename{} for DenseNet-BC) output channels is performed on the input images. For convolutional layers with kernel size 3$\times$3, each side of the inputs is zero-padded by one pixel to keep the feature-map size fixed. We use 1$\times$1 convolution followed by 2$\times$2 average pooling as transition layers between two contiguous dense blocks. At the end of the last dense block, a global average pooling is performed and then a softmax classifier is attached. The feature-map sizes in the three dense blocks are 32$\times$ 32, 16$\times$16, and 8$\times$8, respectively. We experiment with the basic \methodnameshort{} structure with configurations $\{L\!=\!40, k\!=\!12\}$, $\{L\!=\!100,k\!=\!12\}$ and $\{L\!=\!100,k\!=\!24\}$. For \methodnameshort{}-BC, the networks with configurations $\{L\!=\!100, k\!=\!12\}$, $\{L\!=\!250,k\!=\!24\}$ and $\{L\!=\!190,k\!=\!40\}$ are evaluated.

In our experiments on ImageNet, we use a \methodnameshort{}-BC structure with 4 dense blocks on 224$\times$224 input images. The initial convolution layer comprises $2k$ convolutions of size 7$\times$7 with stride 2; the number of feature-maps in all other layers also follow from setting $k$. The exact network configurations we used on ImageNet are shown in Table~\ref{densenet-imagenet}.

%% file: experiments.tex
\section{Experiments}
\input{result/caption}
\input{result/table}

\label{sec:results}
\label{sec:experiment-results}
  We empirically demonstrate \methodnameshort{}'s effectiveness on several benchmark datasets
  and compare with state-of-the-art architectures, especially with ResNet and its variants.

\subsection{Datasets}
\paragraph{CIFAR.}
      The two CIFAR datasets \cite{cifar} consist of colored natural images with 32$\times$32 pixels. CIFAR-10 (C10) consists of images drawn from 10 and CIFAR-100 (C100) from 100 classes. The training and test sets contain 50,000 and 10,000 images respectively, and we hold out 5,000 training images as a validation set.
      We adopt a standard data augmentation scheme (mirroring/shifting) that is widely used for these two datasets \cite{resnet, stochastic, fractalnet, netinnet, fitnet, dsn, allcnn, highway}. We denote this data augmentation scheme by a ``+'' mark at the end of the dataset name (\emph{e.g.}, C10+). For preprocessing, we normalize the data using the channel means and standard deviations. For the final run we use all 50,000 training images and report the final test error at the end of training.

\vspace{-2 ex}
\paragraph{SVHN.}
    The Street View House Numbers (SVHN) dataset \cite{svhn} contains 32$\times$32 colored digit images. There are 73,257 images in the training set, 26,032 images in the test set, and 531,131 images for additional training. Following common practice~\cite{maxout, stochastic, dsn, netinnet, sermanet2012convolutional} we use all the training data without any data augmentation, and a validation set with 6,000 images is split from the training set. We select the model with the lowest validation error during training and report the test error. We follow \cite{wide} and divide the pixel values by 255 so they are in the $[0, 1]$ range.
%

%

 \begin{table*}[ht]
     \vspace{-8pt}
\begin{minipage}[b]{0.35\linewidth}
\centering
\footnotesize
\def\arraystretch{1.8}
\setlength{\tabcolsep}{2pt}
    \begin{tabular}[b]{c|c|c}\hline
      Model & top-1 & top-5 \\ \hline
      \methodnameshort-121& 25.02 / 23.61 & 7.71 / 6.66\\
      \methodnameshort-169& 23.80 / 22.08 & 6.85 / 5.92\\
      \methodnameshort-201& 22.58 / 21.46 & 6.34 / 5.54\\
      \methodnameshort-264& 22.15 / 20.80 & 6.12 / 5.29\\ \hline
    \end{tabular}
    \vspace{8pt}
    \caption{The top-1 and top-5 error rates on the ImageNet validation set, with single-crop / 10-crop testing.}
    \label{table:imagenet-numbers}
\end{minipage}\hfill
\begin{minipage}[b]{0.60\linewidth}
\centering
\includegraphics[width=1.0\textwidth]{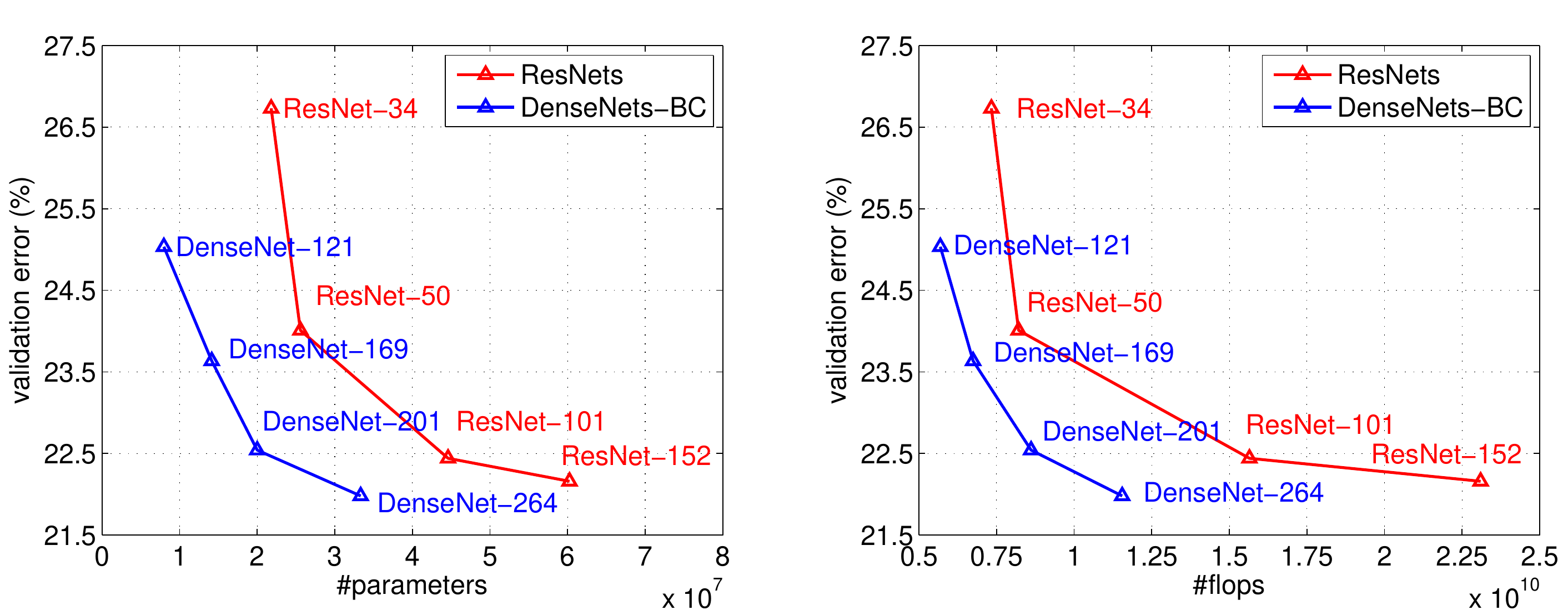}
\captionof{figure}{Comparison of the DenseNets and ResNets top-1 error rates (single-crop testing) on the ImageNet validation dataset as a function of learned parameters (\emph{left}) and FLOPs during test-time (\emph{right}).}
\label{fig:imagenet}
\end{minipage}
\vspace{-2ex}
\end{table*}

\vspace{-2 ex}
\paragraph{ImageNet.} The ILSVRC 2012 classification dataset \cite{deng2009imagenet} consists 1.2 million images for training, and 50,000 for validation, from $1,000$ classes. We adopt the same data augmentation scheme for training images as in \cite{blogresnet,resnet,identity-mappings}, and apply a single-crop or 10-crop with size 224$\times$224 at test time. Following \cite{resnet,identity-mappings,stochastic}, we report classification errors on the validation set.



\subsection{Training}
All the networks are trained using stochastic gradient descent (SGD). On CIFAR and SVHN we train using batch size 64 for 300 and 40 epochs, respectively. The initial learning rate is set to 0.1, and is divided by 10 at 50\% and 75\% of the total number of training epochs. On ImageNet, we train models for 90 epochs with a batch size of 256. The learning rate is set to 0.1 initially, and is lowered by 10 times at epoch 30 and 60. Note that a naive implementation of DenseNet may contain memory inefficiencies. To reduce the memory consumption on GPUs, please refer to our technical report on the memory-efficient implementation of DenseNets \cite{pleiss2017memory}.


Following \cite{blogresnet}, we use a weight decay of $10^{-4}$ and a Nesterov momentum \cite{nesterov} of 0.9 without dampening. We adopt the weight initialization introduced by \cite{init}. For the three datasets without data augmentation, \emph{i.e.}, C10, C100 and SVHN, we add a dropout layer \cite{dropout} after each convolutional layer (except the first one) and set the dropout rate to 0.2. The test errors were only evaluated once for each task and model setting.



\subsection{Classification Results on CIFAR and SVHN}

We train \methodnameshort{}s with different depths, $L$, and \stepsizename{}s, $k$.
The main results on CIFAR and SVHN are shown in Table~\ref{my-label}.
To highlight general trends, we mark all results that outperform the existing state-of-the-art  in {\textbf{boldface}} and the overall best result in {\color{blue} \textbf{blue}}.

\vspace{-2 ex}
\paragraph{Accuracy.}
Possibly the most noticeable trend may originate from the bottom row of Table~\ref{my-label}, which shows that \methodnameshort{}-BC with $L\!=\!190$ and $k\!=\!40$ outperforms the existing state-of-the-art consistently on \emph{all} the CIFAR datasets. Its error rates of 3.46\% on C10+ and 17.18\% on C100+ are significantly lower than the error rates achieved by wide ResNet architecture~\cite{wide}. Our best results on C10 and C100 (without data augmentation) are even more encouraging: both are close to 30\% lower than FractalNet with drop-path regularization \cite{fractalnet}. On SVHN, with dropout, the \methodnameshort{} with $L\!=\!100$ and $k\!=\!24$ also surpasses the current best result achieved by wide ResNet. However, the 250-layer DenseNet-BC doesn't further improve the performance over its shorter counterpart. This may be explained by that SVHN is a relatively easy task, and extremely deep models may overfit to the training set.



\vspace{-2 ex}
\paragraph{Capacity.} Without compression or bottleneck layers, there is a general trend that \methodnameshorts{}  perform better as $L$ and $k$ increase. We attribute this primarily to the corresponding growth in model capacity. This is best demonstrated by the column of C10+ and C100+. On C10+,  the error drops from 5.24\% to 4.10\% and finally to 3.74\% as the number of parameters increases from 1.0M, over 7.0M to 27.2M. On C100+, we observe a similar trend. This suggests that \methodnameshorts{} can utilize the increased representational power of bigger and deeper models. It also indicates that they do not suffer from overfitting or the optimization difficulties of residual networks~\cite{resnet}.

\begin{figure*}[!ht]
\centerline{
\resizebox{\textwidth}{!}{
    \includegraphics[width=0.28\linewidth]{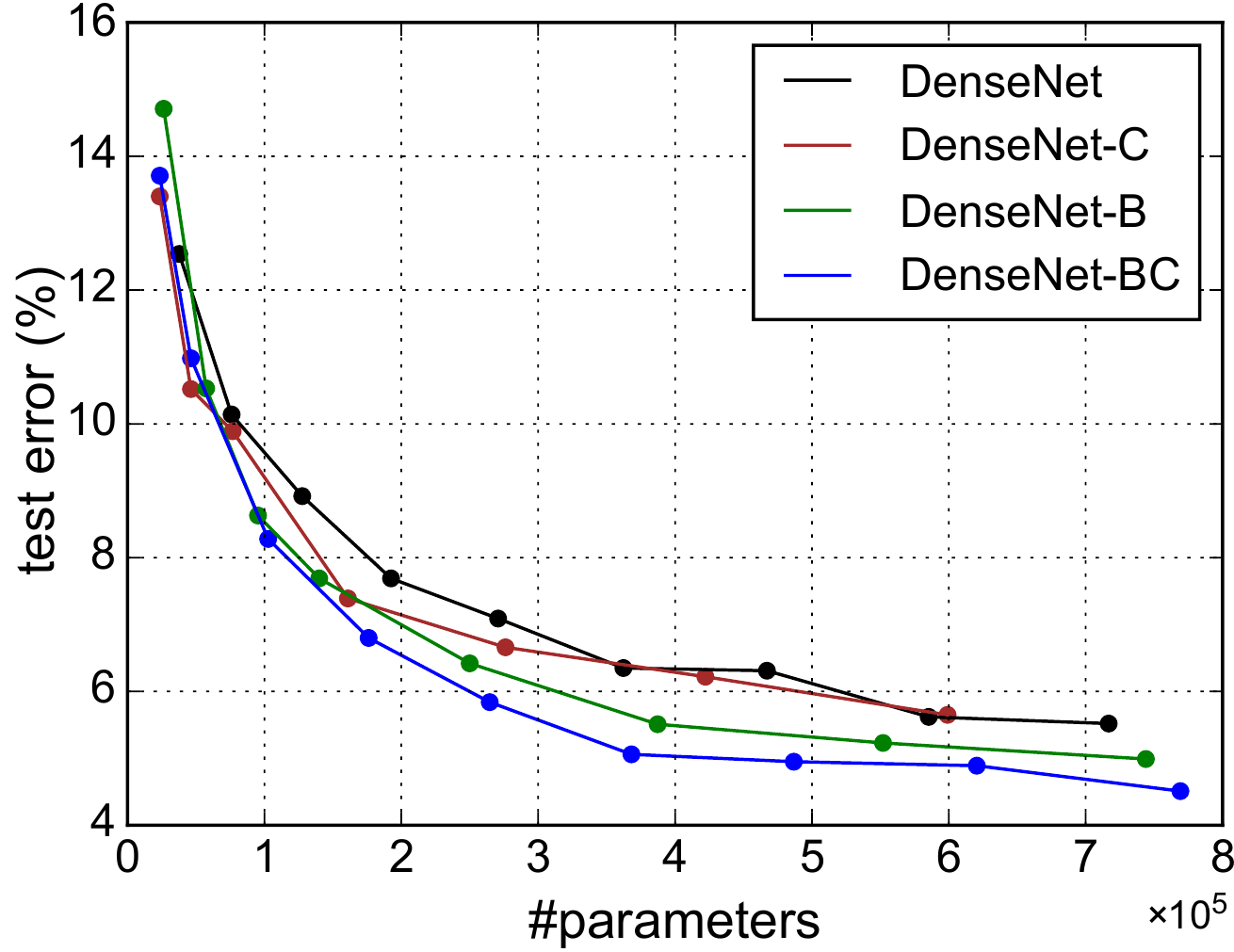}
    \includegraphics[width=0.28\linewidth]{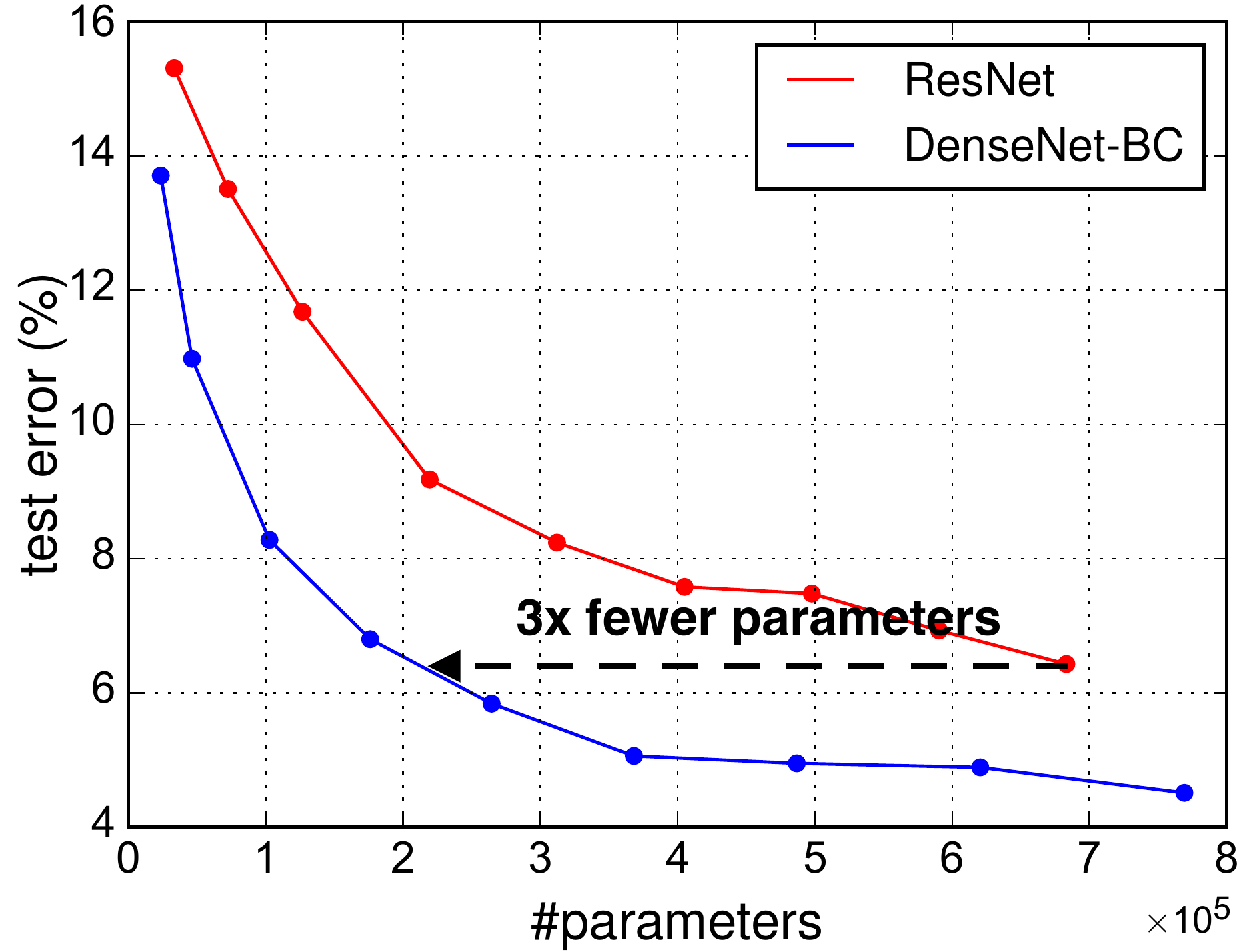}
    \includegraphics[width=0.38\textwidth]{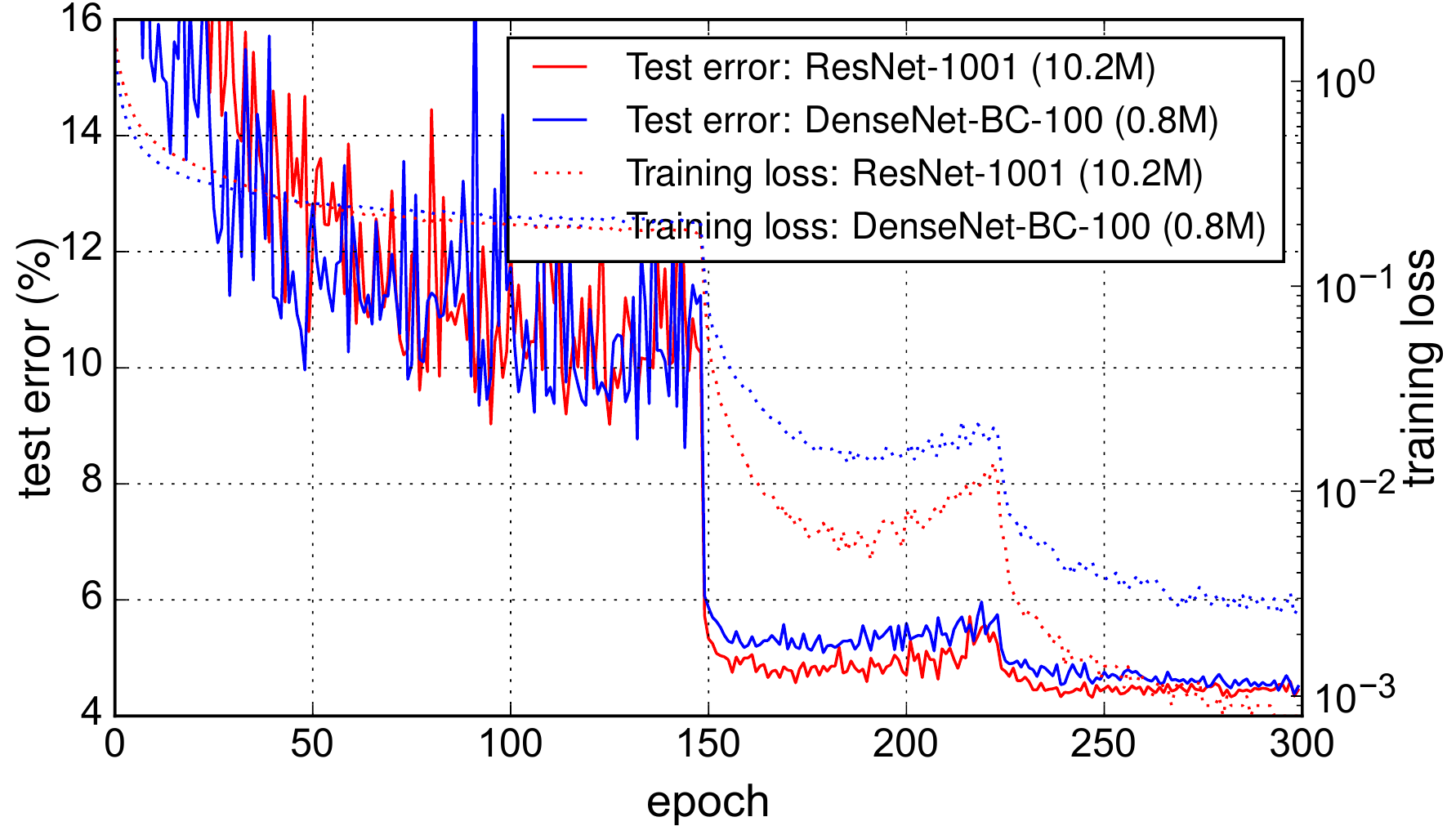}}
    }
          \vspace{-1ex}
      \caption{\emph{Left:} Comparison of the parameter efficiency on C10+ between \methodnameshort{} variations. \emph{Middle:} Comparison of the parameter efficiency between \methodnameshort{}-BC and (pre-activation) ResNets. \methodnameshort{}-BC requires about 1/3 of the parameters as ResNet to achieve comparable accuracy.
      \emph{Right:} Training and testing curves of the 1001-layer pre-activation ResNet \cite{identity-mappings} with more than 10M parameters and a 100-layer \methodnameshort{} with only 0.8M parameters.}
      \label{fig:params}
      \vspace{-3ex}
\end{figure*}

\vspace{-2 ex}
\paragraph{Parameter Efficiency.}
The results in Table~\ref{my-label} indicate that \methodnameshorts{} utilize parameters more efficiently than alternative architectures (in particular, ResNets). The \methodnameshort{}-BC with bottleneck structure and dimension reduction at transition layers is particularly parameter-efficient. For example, our 250-layer model only has 15.3M parameters, but it consistently outperforms other models such as FractalNet and Wide ResNets that have more than 30M parameters. We also highlight that \methodnameshort{}-BC with $L\!=\!100$ and $k\!=\!12$ achieves comparable performance  (\emph{e.g.}, 4.51\% vs 4.62\% error on C10+, 22.27\% vs 22.71\% error on C100+) as the 1001-layer pre-activation ResNet using 90\% fewer parameters.
Figure~\ref{fig:params} (right panel) shows the training loss and test errors of these two networks on C10+. The 1001-layer deep ResNet converges to a lower training loss value but a similar test error. We analyze this effect in more detail below.

\vspace{-2 ex}
\paragraph{Overfitting.}  One positive side-effect of the more efficient use of parameters is a tendency of \methodnameshorts{} to be less prone to overfitting.
We observe that on the datasets without data augmentation, the improvements of \methodnameshort{} architectures over prior work are particularly pronounced. On C10, the improvement denotes a 29\% relative reduction in error from 7.33\% to 5.19\%. On C100, the reduction is about 30\% from 28.20\% to 19.64\%.
In our experiments, we observed potential overfitting in a single setting: on C10, a 4$\times$ growth of parameters produced by increasing $k\!=\!12$ to $k\!=\!24$ lead to a modest increase in error from 5.77\% to 5.83\%.
The \methodnameshort{}-BC bottleneck and compression layers appear to be an effective way to counter this trend.

\subsection{Classification Results on ImageNet}
We evaluate \methodnameshort{}-BC with different depths and growth rates on the ImageNet classification task, and compare it with state-of-the-art ResNet architectures. To ensure a fair comparison between the two architectures, we eliminate all other factors such as differences in data preprocessing and optimization settings by adopting the publicly available Torch implementation for ResNet by \cite{blogresnet}\footnote{\url{https://github.com/facebook/fb.resnet.torch}}.
We simply replace the ResNet model with the \methodnameshort{}-BC network, and keep all the experiment settings \emph{exactly} the same as those used for ResNet.

We report the single-crop and 10-crop validation errors of \methodnameshorts{} on ImageNet in Table~\ref{table:imagenet-numbers}. Figure~\ref{fig:imagenet} shows the single-crop top-1 validation errors of \methodnameshorts{} and ResNets as a function of the number of parameters (left) and FLOPs (right). The results presented in the figure reveal that \methodnameshorts{} perform on par with the state-of-the-art ResNets, whilst requiring significantly fewer parameters and computation to achieve comparable performance. For example, a DenseNet-201 with 20M parameters model yields similar validation error as a 101-layer ResNet with more than 40M parameters. Similar trends can be observed from the right panel, which plots the validation error as a function of the number of FLOPs: a \methodnameshort{} that requires as much computation as a ResNet-50 performs on par with a ResNet-101, which requires twice as much computation.

It is worth noting that our experimental setup implies that we use hyperparameter settings that are optimized for ResNets but not for \methodnameshorts{}. It is conceivable that more extensive hyper-parameter searches may further improve the performance of \methodnameshort{} on ImageNet.


%% file: result/caption.tex
\newcommand{\resultcaption}{Error rates (\%) on CIFAR and SVHN datasets. $k$ denotes network's \stepsizename{}. Results that surpass all competing methods are {\textbf{bold}} and the overall best results are {\color{blue}\textbf{blue}}. ``+'' indicates standard data augmentation (translation and/or mirroring). $\ast$ indicates results run by ourselves. All the results of \methodnameshort{}s without data augmentation (C10, C100, SVHN) are obtained using Dropout. \methodnameshort{}s achieve lower error rates while using fewer parameters than ResNet. 
Without data augmentation, \methodnameshort{} performs better by a large margin.
}

%% file: result/table.tex
\renewcommand{\arraystretch}{1.06}
\setlength{\tabcolsep}{1.0em}
\begin{table*}[]
\centering
\resizebox{0.9\textwidth}{!}{%
\begin{tabular}{l|cc|cc|cc|c}
\hline
\multicolumn{1}{c|}{Method} & \multicolumn{1}{l}{Depth} & \multicolumn{1}{l|}{Params} & C10 & C10+ & C100 & C100+ & SVHN \\ \hline
Network in Network \cite{netinnet} & - & - & 10.41 & 8.81 & 35.68 & - & 2.35 \\
All-CNN \cite{allcnn} & - & - & 9.08 & 7.25 & - & 33.71 & - \\
Deeply Supervised Net \cite{dsn} & - & - & 9.69 & 7.97 & - & 34.57 & 1.92 \\
Highway Network \cite{highway} & - & - & - & 7.72 & - & 32.39 & - \\ \hline
FractalNet \cite{fractalnet} & 21 & 38.6M & 10.18 & 5.22 & 35.34 & 23.30 & 2.01 \\
with Dropout/Drop-path & 21 & 38.6M & 7.33 & 4.60 & 28.20 & 23.73 & 1.87 \\ \hline
ResNet \cite{resnet} & 110 & 1.7M & - & 6.61 & - & - & - \\ \hline
ResNet (reported by \cite{stochastic}) & 110 & 1.7M & 13.63 & 6.41 & 44.74 & 27.22 & 2.01 \\ \hline
ResNet with Stochastic Depth \cite{stochastic} & 110 & 1.7M & 11.66 & 5.23 & 37.80 & 24.58 & 1.75 \\
 & 1202 & 10.2M & - & 4.91 & - & - & - \\ \hline
Wide ResNet \cite{wide} & 16 & 11.0M & - & 4.81 & - & 22.07 & - \\
 & 28 & 36.5M & - & 4.17 & - & 20.50 & - \\
with Dropout & 16 & 2.7M & - & - & - & - & 1.64 \\ \hline
ResNet (pre-activation) \cite{identity-mappings} & 164 & 1.7M & 11.26$^\ast$ & 5.46 & 35.58$^\ast$ & 24.33 & - \\
\multicolumn{1}{c|}{} & 1001 & 10.2M & 10.56$^\ast$ & 4.62 & 33.47$^\ast$ & 22.71 & - \\ \hline
\methodnameshort{} $(k=12)$ & 40 & 1.0M & \textbf{7.00} & 5.24 & \textbf{27.55} & 24.42 & 1.79 \\
\methodnameshort{} $(k=12)$ & 100 & 7.0M & \textbf{5.77} & \textbf{4.10} & \textbf{23.79} & \textbf{20.20} & 1.67 \\
\methodnameshort{} $(k=24)$ & 100 & 27.2M & \textbf{5.83} & \textbf{3.74} & \textbf{23.42} & \textbf{19.25} & {\color{blue}\textbf{1.59}} \\ \hline
\methodnameshort{}-BC $(k=12)$ & 100 & 0.8M & \textbf{5.92} & 4.51 & \textbf{24.15} & 22.27 & 1.76 \\
\methodnameshort{}-BC $(k=24)$ & 250 & 15.3M & {\color{blue}\textbf{5.19}} & \textbf{3.62} & {\color{blue}\textbf{19.64}} & \textbf{17.60} & 1.74 \\
\methodnameshort{}-BC $(k=40)$ & 190 & 25.6M & - & {\color{blue}\textbf{3.46}} & - & {\color{blue}\textbf{17.18}} & - \\ \hline
\end{tabular}%
}
\vspace{0.5ex}
\caption{\resultcaption}
\label{my-label}
\vspace{-1ex}
\end{table*} 

%% file: analytic.tex
\section{Discussion}

Superficially, \methodnameshort{}s are quite similar to ResNets: Eq. (\ref{eqn:densenet}) differs from Eq.~(\ref{eq:resnet}) only in that the inputs to $H_\ell(\cdot)$ are concatenated instead of summed. However, the implications of this seemingly small modification lead to substantially different behaviors of the two network architectures.

\vspace{-2ex}
\paragraph{Model compactness.}
As a direct consequence of the input concatenation, the feature-maps learned by any of the \methodnameshort{} layers can be accessed by all subsequent layers. This encourages feature reuse throughout the network, and leads to more compact models.

The left two plots in Figure~\ref{fig:params} show the result of an experiment that aims to compare the parameter efficiency of all variants of \methodnameshorts{} (left) and also a comparable ResNet architecture (middle).
We train multiple small networks with varying depths on C10+ and plot their test accuracies as a function of network parameters.
In comparison with other popular network architectures, such as AlexNet \cite{alexnet} or VGG-net \cite{vgg}, ResNets with pre-activation use fewer parameters while typically achieving better results~\cite{identity-mappings}. Hence, we compare \methodnameshort{} ($k=12$) against this architecture. The training setting for \methodnameshort{} is kept the same as in the previous section.

The graph shows that \methodnameshort{}-BC is consistently the most parameter efficient variant of \methodnameshort{}.
Further, to achieve the same level of accuracy, \methodnameshort{}-BC only requires around 1/3 of the parameters of ResNets (middle plot). This result is in line with the results on ImageNet we presented in Figure~\ref{fig:imagenet}. The right plot in Figure~\ref{fig:params} shows that a DenseNet-BC with only 0.8M trainable parameters is able to achieve comparable accuracy as the 1001-layer (pre-activation) ResNet \cite{identity-mappings} with 10.2M parameters.


\vspace{-2ex}
\paragraph{Implicit Deep Supervision.}
One explanation for the improved accuracy of \methodname{}s may be that individual layers receive additional supervision from the loss function through the shorter connections. One can interpret \methodnameshorts{} to perform a kind of ``deep supervision''. The benefits of deep supervision have previously been shown in deeply-supervised nets (DSN; \cite{dsn}), which have classifiers attached to every hidden layer, enforcing the intermediate layers to learn discriminative features.

\methodnameshorts{} perform a similar deep supervision in an implicit fashion: a single classifier on top of the network provides direct supervision to all layers through at most two or three transition layers. However, the loss function and gradient of \methodnameshorts{} are substantially less complicated, as the same loss function is shared between all layers.

\vspace{-2ex}
\paragraph{Stochastic vs. deterministic connection.} There is an interesting connection between \methodname{}s and stochastic depth regularization of residual networks~\cite{stochastic}. In stochastic depth, layers in residual networks are randomly dropped, which creates direct connections between the surrounding layers. As the pooling layers are never dropped, the network results in a similar connectivity pattern as \methodnameshort{}: there is a small probability for any two layers, between the same pooling layers, to be directly connected---if all intermediate layers are randomly dropped.  Although the methods are ultimately quite different, the \methodnameshort{} interpretation of stochastic depth may provide insights into the success of this regularizer.

\begin{figure}[t]
      \centering
      \includegraphics[width=0.48 \textwidth]{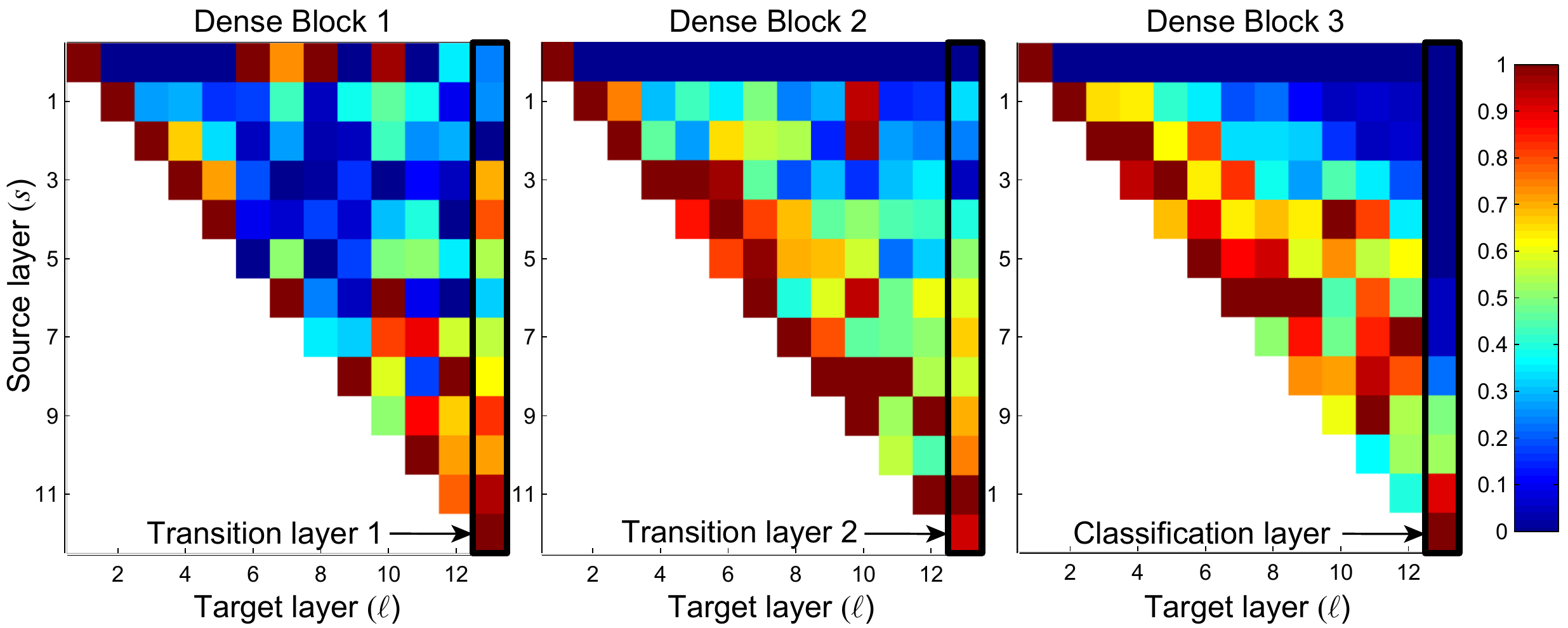}
            \vspace{-2ex}
      \caption{The average absolute filter weights of convolutional layers in a trained \methodnameshort{}. The color of pixel $(s,\ell)$ encodes the average $L1$ norm (normalized by number of input feature-maps) of the weights connecting convolutional layer $s$ to $\ell$ within a dense block. Three columns highlighted by black rectangles correspond to two transition layers and the classification layer. The first row encodes weights connected to the input layer of the dense block. }
      \label{fig:weight}
      \vspace{-2ex}
\end{figure}

\paragraph{Feature Reuse.}
By design, \methodnameshorts{} allow layers access to feature-maps from all of its preceding layers (although sometimes through transition layers). We conduct an experiment to investigate if  a trained network takes advantage of this opportunity.
We first train a \methodnameshort{} on C10+ with $L\!=\!40$ and $k\!=\!12$. For each convolutional layer $\ell$ within a block, we compute the average (absolute) weight assigned to connections with layer $s$. Figure~\ref{fig:weight} shows a heat-map for all three dense blocks.
The average absolute weight serves as a surrogate for the dependency of a convolutional layer on its preceding layers.
A red dot in position ($\ell,s$) indicates that the layer $\ell$  makes, on average, strong use of feature-maps produced $s$-layers before.
Several observations can be made from the plot:
\vspace{-2pt}
\begin{enumerate}
\setlength{\itemsep}{-2.6pt}
\item All layers spread their weights over many inputs within the same block. This indicates that features extracted by very early layers are, indeed, directly used by deep layers throughout the same dense block.
\item The weights of the transition layers also spread their weight across all layers within the preceding dense block, indicating information flow from the first to the last layers of the \methodnameshort{} through few indirections.
\item The layers within the second and third dense block consistently assign the least weight to the outputs of the transition layer (the top row of the triangles), indicating that the transition layer outputs many redundant features (with low weight on average).  This is in keeping with the strong results of \methodnameshort{}-BC where exactly these outputs are compressed.
\item Although the final classification layer, shown on the very right, also uses weights across the entire dense block, there seems to be a concentration towards final feature-maps, suggesting that there may be some more high-level features produced late in the network.
\end{enumerate}
%

%% file: conclusion.tex
\section{Conclusion}
We proposed a new convolutional network architecture, which we refer to as \methodnamecap{} (\methodnameshort{}). It introduces direct connections between any two layers with the same feature-map size.
We showed that \methodnameshorts{} scale naturally to hundreds of layers, while exhibiting no optimization difficulties. In our experiments, \methodnameshorts{} tend to yield consistent improvement in accuracy with growing number of parameters, without any signs of performance degradation or overfitting. Under multiple settings, it achieved state-of-the-art results across several highly competitive datasets. Moreover, \methodnameshorts{} require substantially fewer parameters and less computation to achieve state-of-the-art performances.
Because we adopted hyperparameter settings optimized for residual networks in our study, we believe that further gains in accuracy of \methodnameshorts{} may be obtained by more detailed tuning of hyperparameters and learning rate schedules. 

Whilst following a simple connectivity rule, \methodnameshorts{} naturally integrate the properties of identity mappings, deep supervision, and diversified depth. They allow feature reuse throughout the networks and can consequently learn more compact and, according to our experiments, more accurate models. Because of their compact internal representations and reduced feature redundancy, \methodnameshorts{} may be good feature extractors for various computer vision tasks that build on convolutional features, \emph{e.g.}, ~\cite{gardner2015deep,gatys2015neural}. We plan to study such feature transfer with \methodnameshorts{} in future work.